\documentclass[10pt,runningheads]{llncs}

\usepackage{savesym}
\usepackage{amsmath}
\savesymbol{iint}
\usepackage{txfonts}
\usepackage{times}
\restoresymbol{TXF}{iint}
\usepackage{multirow} 
\usepackage{booktabs} 

\usepackage{url}

\usepackage[utf8x]{inputenc}

\usepackage{tikz}
\usepackage{pgfplots}
\usetikzlibrary{arrows,decorations.pathmorphing,backgrounds,positioning,fit,plotmarks,shadows}
\pgfplotsset{compat=1.3,every axis/.append style={font=\scriptsize}, every legend/.append style={font=\scriptsize}}
\usepackage{color}

\tikzstyle{fancylabel} = [text=white, inner sep=0pt, minimum size=15pt, yshift=6pt, xshift=1pt]

\newcommand{\EEqref}[1]{\text{Eq.}~\eqref{#1}}

\newcommand{\ETAL}{et al.}

\usepackage{amsmath}

\newcommand{\rmd}{\mathrm{d}}  

\newcommand{\BF}[1]{\ensuremath{\textbf{#1}}}
\newcommand{\q}{\ensuremath{\mathbf{q}}}
\newcommand{\rr}{\ensuremath{\mathbf{r}}}
\newcommand{\RR}{\ensuremath{\mathbf{R}}}

\newcommand{\uu}{\ensuremath{\mathbf{u}}}

\newcommand{\mathbi}[1]{\ensuremath{\textbf{\textit{#1}}}}
\newcommand{\VV}[1]{\ensuremath{\textbf{\textit{#1}}}}

\newcommand{\Vc}{\mathbi{c}}

\newcommand{\Va}{\mathbi{a}}
\newcommand{\Ve}{\mathbi{e}}

\newcommand{\BB}[1]{\boldsymbol{#1}}

\usepackage[bookmarks=true, 
            final=true,
            hyperfootnotes=true,
            ]{hyperref}

\usepackage{graphicx}
\usepackage{grffile} 
\graphicspath{{pics/}{figs/}}

\newcommand{\RMarginChange}[1]{}
\newcommand{\RMarginComment}[1]{}

\setkeys{Gin}{width=1.0\textwidth}

\emergencystretch=\hsize
\begin{document}
\frontmatter          
\pagestyle{headings}  
\mainmatter              

\title{Regularized Spherical Polar Fourier Diffusion MRI with Optimal Dictionary Learning}
\titlerunning{Regularized SPFI with Optimal Dictionary Learning}

\author{ Jian Cheng\inst{1}, Tianzi Jiang\inst{2}, Rachid Deriche\inst{3},  Dinggang Shen\inst{1}, Pew-Thian Yap\inst{1}    }

\institute{ 
  Department of Radiology and BRIC, The University of North Carolina at Chapel Hill, USA 
  \and 
  CCM, LIAMA, Institute of Automation, Chinese Academy of Sciences, China 
  \and
  Athena Project Team, INRIA Sophia Antipolis, France 
  \newline \email{\{jian\_cheng,dgshen,ptyap\}@med.unc.edu} 
}
\authorrunning{J. Cheng, T. Jiang, R. Deriche, D. Shen, P.-T. Yap}   


\maketitle              

\begin{abstract}
Compressed Sensing (CS) takes advantage of signal sparsity or compressibility and allows superb signal reconstruction from relatively few measurements. 
Based on CS theory, a suitable dictionary for sparse representation of the signal is required. 
In diffusion MRI (dMRI), CS methods were proposed to reconstruct diffusion-weighted signal and the Ensemble Average Propagator (EAP), 
and there are two kinds of Dictionary Learning (DL) methods: 
1) Discrete Representation DL (DR-DL), and 2) Continuous Representation DL (CR-DL). 
DR-DL is susceptible to numerical inaccuracy owing to interpolation and regridding errors in a discretized $q$-space. 
%
In this paper, we propose a novel CR-DL approach, called Dictionary Learning - Spherical Polar Fourier Imaging (DL-SPFI) for effective compressed-sensing reconstruction of the $\q$-space diffusion-weighted signal and the EAP. 
In DL-SPFI, an dictionary that sparsifies the signal is learned from the space of continuous Gaussian diffusion signals. 
The learned dictionary is then adaptively applied to different voxels using a weighted LASSO 
framework for robust signal reconstruction. 
The adaptive dictionary is proved to be optimal. 
%
%
Compared with the start-of-the-art CR-DL and DR-DL methods proposed by Merlet \ETAL\ and Bilgic \ETAL, respectively, our work offers the following advantages. 
First, the learned dictionary is proved to be optimal for Gaussian diffusion signals. 
Second, to our knowledge, this is the first work to learn a voxel-adaptive dictionary. 
The importance of the adaptive dictionary in EAP reconstruction will be demonstrated theoretically and empirically.
Third, optimization in DL-SPFI is only performed in a small subspace resided by the SPF coefficients, as opposed to the $\q$-space approach utilized by Merlet {\ETAL} 
We experimentally evaluated DL-SPFI with respect to L1-norm regularized SPFI (L1-SPFI), which uses the original SPF basis, and the DR-DL method proposed by Bilgic {\ETAL} 
The experiment results on synthetic and real data indicate that the learned dictionary produces sparser coefficients than the original SPF basis 
and results in significantly lower reconstruction error than Bilgic {\ETAL}'s method. 


\end{abstract}

\section{Introduction}
\label{sec:intro}

Diffusion MRI (dMRI) is a unique non-invasive technique for investigation of white matter microstructure in the human brain. 
A central problem in dMRI is to estimate the Ensemble Average Propagator (EAP) $P(\RR)$, which describes fully the probability distribution of water molecule displacement $\RR$, from a limited number of measurements of the signal attenuation $E(\q)$ in the $\q$ (wave-vector) space.
Under narrow pulse condition, $E(\q)$ and $P(\RR)$ are related by the Fourier transform, i.e.,
$P(\RR) = \int_{\mathbb{R}^3} E(\q) \exp(-2\pi i \q^T\RR)\rmd \q$. 
Various methods have been proposed for reconstructing the EAP. The most common method is Diffusion Tensor Imaging (DTI).
However, due to its Gaussian assumption, DTI is incapable of modeling complex non-Gaussian diffusion resulting from crossing fibers. 
Diffusion Spectrum Imaging (DSI) acquires measurements for more than 500 discrete points in the $\q$-space 
and performs Fourier transform numerically to obtain the EAP, followed by an numerical radial integration to estimate the Orientation Distribution Function (ODF)~\cite{Wedeen2005}. However, the long scanning time ($\approx$ 1 hour) required by DSI significantly limits its utility, especially in clinical settings.
Spherical Polar Fourier Imaging (SPFI), by leveraging a continuous representation of $E(\q)$, requires a more moderate number of signal measurements. This continuous representation is based on the Spherical Polar Fourier (SPF) basis and provides closed-form expressions for EAP and ODF computation~\cite{AssemlalMIA2009,Cheng_PDF_MICCAI2010}. 

Recovering a latent function from a small number of samples in Fourier domain is a 
classic problem in Compressed Sensing (CS) theory~\cite{donoho2006compressed}, 
where a good basis that allows sparse representation is crucial for the reconstruction. 
Although some analytic bases, including discrete basis like wavelets~\cite{menzel_MRM11} and continuous basis like the SPF basis,
have been proposed as sparse bases for EAP estimation, 
based on CS theory, a sparser basis can be learned from well chosen exemplars via Dictionary Learning (DL) techniques~\cite{aharon_KSVD_TSP06,mairal_JMLR10}. 
Bilgic \ETAL~\cite{bilgic_MRM12} learns a discrete dictionary via the K-SVD~\cite{aharon_KSVD_TSP06} approach and uses it in FOCal Underdetermined System Solver (FOCUSS) algorithm for EAP estimation. 
This strategy dramatically reduces the number of samples and scanning time required by DSI. 
However, Bilgic \ETAL's approach suffers from numerical errors similar to DSI because their dictionary is composed of a set of discrete basis vectors. 
On the other hand, 
Merlet \ETAL~\cite{merlet_dictionary_MICCAI12} learns a continuous dictionary, 
parametrized as a linear combination of some atoms adopted from SPF basis, from synthetic Gaussian signals, 
where the learned basis has the closed forms for ODF and EAP estimation due to the results of SPF basis~\cite{Cheng_PDF_MICCAI2010}. 
However, there are some inherent limitations in both theoretical
analysis and practical usage in~\cite{merlet_dictionary_MICCAI12}. 
For example, they learned the scale parameter $\zeta$ associated with the SPF basis from the training data, instead of the testing data. 
We shall show in the current paper that the optimal scale $\zeta$ should be adaptively estimated from testing data.  
In addition, they have also neglected isotropic exemplars in the training data, causing over-fitting problems in less anisotropic areas such as the grey matter.


In this paper, we propose a novel CR-DL approach, called Dictionary Learning - Spherical Polar Fourier Imaging (DL-SPFI), for effective compressed-sensing reconstruction of the diffusion signal and the EAP. 
Our approach offers a number of advantages over~\cite{merlet_dictionary_MICCAI12}.
First, we dramatically reduce the dimensionality of the optimization problem 
by working in a small subspace of the SPF coefficients, instead of $\q$-space as done in~\cite{merlet_dictionary_MICCAI12}. 
Second, the dictionary learned using our approach can be applied optimally and adaptively to each voxel by voxel-dependent determination of the optimal scale parameter. 
In contrast, both~\cite{merlet_dictionary_MICCAI12} and~\cite{bilgic_MRM12} do not consider inter-voxel variation. 
Third, 
we consider the constraint $E(0)=1$ during both learning and estimation processes. 
Section~\ref{sec:SPFI} provides a brief overview on SPFI and shows how the constraint $E(0)=1$ can be incorporated in SPFI. 
Section~\ref{sec:DL} demonstrates the equivalence between dictionary learning and estimation regularization 
and provides details on DL-SPFI. 
Section~\ref{sec:exp} validates DL-SPFI in comparison with L1-SPFI using the original SPF basis and
FOCUSS with/without DL, as implemented in~\cite{bilgic_MRM12}.

\section{Spherical Polar Fourier Imaging (SPFI) Revisited}
\label{sec:SPFI}

The SPF basis is a continuous complete basis that can represent sparsely Gaussian-like signals~\cite{AssemlalMIA2009,Cheng_PDF_MICCAI2010}. 
If $E(\q)$ is represented by the SPF basis $\{B_{nlm}(\q)\}=\{G_n(q) Y_l^m(\uu)\}$, i.e., 
\begin{footnotesize}
\begin{equation}\label{eq:SPF} 
  E(q\uu) = \sum_{n=0}^N\sum_{l=0}^L\sum_{m=-l}^l a_{nlm}B_{nlm}(\q), \quad B_{nlm}(\q) = G_n(q|\zeta) Y_l^m(\uu)
\end{equation} 
\end{footnotesize}%
where $G_n(q|\zeta)=\left[\frac{2}{\zeta ^{3/2}} \frac{n!}{\Gamma(n+3/2)}\right ]^{1/2}\exp\left(-\frac{q^2}{2\zeta}\right) L_n^{1/2}(\frac{q^2}{\zeta})$ is the Gaussian-Laguerre polynomial, $Y_l^m(\uu)$ is the real Spherical Harmonic (SH) basis, 
and $\zeta$ is the scale parameter, 
then the EAP $P(\RR)$  is represented by the Fourier dual SPF (dSPF) basis in~\EEqref{eq:dSPF}, where 
$B^{\text{dual}}_{nlm}(\RR)$ was proved to be the Fourier transform of $B_{nlm}(\q)$. 
The definition of the dSPF basis can be found in~\cite{Cheng_PDF_MICCAI2010}. 
\begin{footnotesize}
\begin{equation} \label{eq:dSPF}
P(R\rr)= \sum_{n=0}^N\sum_{l=0}^L\sum_{m=-l}^l a_{nlm}F_{nl}(R)Y_l^m(\rr)
\qquad  
B^{\text{dual}}_{nlm}(\RR) = F_{nl}(R) Y_l^m(\rr)
\end{equation}
\end{footnotesize}%
%

The SPF coefficients $\Va=(a_{000},\dots,a_{NLL})^T$ can be estimated from the signal attenuation measurements $\{E_i\}$ 
via least square fitting with $l_{2}$-norm or $l_{1}$-norm regularization, where the constraint $E(0)=1$ can be imposed by adding artificial samples at $q=0$~\cite{Cheng_PDF_MICCAI2010,cheng_L1SPFI_CDMRI2011}. 
Here we propose an alternative continuous approach to impose this constraint. 
From $E(0)=1$, we have $\sum_{0}^N a_{nlm}G_n(0)=\sqrt{4\pi} \delta_l^0$, $0\leq l\leq L$, $-l\leq m\leq l$. Based on this, we can separate the coefficient vector $\Va$ into $\Va= (\Va_0^T, \Va'^T)^T$, 
where $\Va_0=(a_{000},\dots,a_{0LL})^T$, $\Va'=(a_{100},\dots,a_{NLL})^T$, 
and represent $\Va_0$ using $\Va'$, i.e.,
\begin{footnotesize}
  \begin{equation}\label{eq:a0}
    a_{0lm} = \frac{1}{G_0(0)} \left( \sqrt{4\pi}\delta_l^0 -\sum_{n=1}^{N} a_{nlm}G_n(0) \right), \quad 0\leq l \leq L, \quad -l\leq m \leq l 
  \end{equation}
\end{footnotesize}%
Based on~\EEqref{eq:SPF}, the $l_{1}$-norm regularized estimation of $\Va'$, called $l_{1}$-SPFI~\cite{cheng_L1SPFI_CDMRI2011}, can be formulated as 
\begin{footnotesize}
\begin{equation}
 \min_{\Va'} \| \BF{M}'\Va' -\VV{e}' \|_2^2 +  \| \boldsymbol\Lambda \Va' \|_{1} 
\end{equation}
\end{footnotesize}%
\begin{footnotesize}
\begin{equation}\label{eq:SPFI:LS_E0_M}
  \BF{M}'= \left[  
\begin{smallmatrix}  
  \left( G_1(q_1|\zeta) -\frac{G_1(0|\zeta)}{G_0(0|\zeta)}G_0(q_1|\zeta) \right)Y_0^0(\uu_1)    & \cdots   & \left( G_{N}(q_1|\zeta) -\frac{G_{N}(0|\zeta)}{G_0(0|\zeta)}G_0(q_1|\zeta) \right)Y_L^L(\uu_1)\\     
  \vdots           & \ddots  & \vdots \\  
  \left( G_1(q_{S}|\zeta) -\frac{G_1(0|\zeta)}{G_0(0|\zeta)}G_0(q_{S}|\zeta) \right)Y_0^0(\uu_{S}) & \cdots  & \left( G_{N}(q_{S}|\zeta) -\frac{G_{N}(0|\zeta)}{G_0(0|\zeta)}G_0(q_{S}|\zeta) \right)Y_L^L(\uu_{S}) 
\end{smallmatrix}
\right], 
~
  \VV{e}'=
  \left[ 
\begin{smallmatrix}  
  E_1-\frac{G_0(q_1)}{G_0(0)}\\
  \vdots\\
  E_{S}-\frac{G_0(q_{S})}{G_0(0)}
\end{smallmatrix}
\right],
\end{equation}
\end{footnotesize}%
where $\|\cdot \|_{p}$ denotes the $l_{p}$-norm, $\{E_i\}_{i=1}^S$ are the $S$ signal attenuation measurements in $\q$-space, 
and the regularization matrix $\boldsymbol\Lambda$ can
be devised as a diagonal matrix with elements $\Lambda_{nlm}=\lambda_l l^2(l+1)^2+\lambda_n n^2(n+1)^2$ 
to sparsify the coefficients, 
where $\lambda_l$ and $\lambda_n$ are the regularization parameters for the angular and radial components. 
Note that $E(\q)-\frac{G_0(q)}{G_0(0)}=E(\q)-\exp(-\frac{q^2}{2\zeta})$ is the signal removing the approximated isotropic Gaussian part, 
and $\left( G_n(q)-\frac{G_n(0)}{G_0(0)} G_0(q) \right)Y_l^m(\uu)$ is the basis $G_n(q)Y_l^m(\uu)$ removing the isotropic Gaussian part. 
After estimating $\Va'$, $\Va_0$ can be obtained using~\EEqref{eq:a0}, and the estimated EAP $\Va$ satisfies $E(0)=1$. 

\section{Dictionary Learning and Regularization}
\label{sec:DL}

\noindent\textbf{Equivalence Between Dictionary Learning and Regularization Design}. 
It was shown in \cite{cheng_L1SPFI_CDMRI2011} that a well-designed regularization matrix enhances coefficient sparsity for better reconstruction.  
A better regularization matrix can be learned from a set of given signals $\{\Ve_i'\}$. 
In fact, learning the regularization matrix from data is equivalent to the so-called dictionary learning, i.e.,
\begin{small}
\begin{equation}\label{eq:DL}
\underbrace{\min_{\BF{A}', \boldsymbol\Lambda, \zeta } \sum_i \| \boldsymbol\Lambda \Va_i' \|_1~\text{s.t.}~\| \BF{M}'\Va_j' -\Ve_j' \|_2 \leq \epsilon_{\text{DL}},~\forall j}_{\text{Regularization Design}}
~  \leftrightarrow ~   
  \underbrace{\min_{\BF{C}, \BF{D},\zeta } \sum_{i} \| \Vc_i \|_1~\text{s.t.}~\| \BF{M}'\BF{D} \Vc_j -\Ve_j' \|_2 \leq \epsilon_{\text{DL}},~\forall j}_{\text{Dictionary Learning}}
\end{equation}
\end{small}%
where $\BF{A}=(\Va'_1,\dots,\Va'_Q)$ is the SPF coefficient matrix. 
The transform matrix $\BF{D}$ will result in a transformed SPF basis $\BF{M}'\BF{D}$ that can be used for even sparser representation of the signal.
 $\BF{C}=(\Vc_1,\dots,\Vc_Q)$ is the new coefficient matrix in association with the transformed basis. 
Here, we include the scale parameter $\zeta$ of the SPF basis as a parameter to be learned. More discussion on this in the next section.
Our formulation is more general than the formulation in~\cite{merlet_dictionary_MICCAI12} for two reasons. 
First, it can be proved that all atoms in the dictionary used in~\cite{merlet_dictionary_MICCAI12} can be represented as a finite linear combination of the SPF basis used in~\EEqref{eq:DL}; the opposite, however, is not true.
\footnote{All proofs in this paper are omitted due to space limitation, available upon request.}
Hence, the space spanned by the atoms in~\cite{merlet_dictionary_MICCAI12} is just a subspace of the space spanned by the SPF basis.
Second, based on the equivalence in~\EEqref{eq:DL}, 
we can further devise a regularization matrix after DL to weight the atoms differently for more effective reconstruction. 

\medskip

\noindent\textbf{Efficient, Optimal, and Adaptive Dictionary Learning}. 
Although it is possible to learn a dictionary from real data, as done in DL-FOCUSS~\cite{bilgic_MRM12}, 
the learned dictionary may be significantly affected by noise and the small sample size. 
An alternative solution to this is to perform DL using some synthetic data that 
approximate well the real signal. 
Similar to~\cite{merlet_dictionary_MICCAI12}, we propose to learn a continuous basis using mixtures of Gaussian signals. Compared with the DL strategy in~\cite{merlet_dictionary_MICCAI12}, our method introduces several theoretical improvements. 
\textbf{1)} Instead of using the DL formulation in \EEqref{eq:DL}, we propose to solve
\begin{footnotesize}
  \begin{equation}\label{eq:DL_final}
    \min_{\BF{C}, \BF{D},\zeta } \sum_{i} \| \Vc_i \|_1~\text{s.t.}~\| \BF{D} \Vc_j -\Va_j' \|_2 \leq \epsilon_{\text{DL}},~\forall j,
  \end{equation}
\end{footnotesize}%
which, due to the orthogonality of the SPF basis, is equivalent to~\EEqref{eq:DL} if $N$ and $L$ are large enough. 
In~\cite{merlet_dictionary_MICCAI12} $\{\VV{e}_i\}$ was generated using thousands of samples, resulting in a high-dimensional minimization problem. 
In contrast, \EEqref{eq:DL_final} works in the small subspace resided by the SPF coefficients and hence significantly reduces the complexity of the learning problem. 
Note that $\zeta$ in~\EEqref{eq:DL_final} is contained inside $\{\Va_j'\}$. 
\textbf{2)} In~\cite{merlet_dictionary_MICCAI12}, a constraint was placed on the sparsity term $\|\Vc_i\|_{1}$, instead of the fitting error term, as is done \EEqref{eq:DL_final}. 
Since there is no prior knowledge on the level of sparsity, it is better to place the constraint on the fitting error. Threshold $\epsilon_{\text{DL}}$ can be chosen simply as $0.01$ for unit-norm normalized $\{\Va_j'\}$.
\textbf{3)} It is not necessary to generate a large sample of signals randomly from the mixture of tensor models, like what is done in~\cite{merlet_dictionary_MICCAI12}. 
We proved in Theorem~\ref{thm:mixture} that the single tensor model is sufficient to learn a dictionary which sparsifies the multi-Gaussian signals. 
That is, the training data $\{\Ve_j\}$ can be generated simply from $\{E(\q|\BB{T})=\exp(-4\pi^2\tau q^2 \uu^T \BB{T} \uu ) \ |\  \BB{T}\in \text{Sym}_+^3 \}$, 
where $\text{Sum}_+^3$ is the space of $3\times 3$ positive symmetric matrices, and $\tau$ is the diffusion time. 
\textbf{4)} 
Compared with the classical DL described in~\cite{aharon_KSVD_TSP06,mairal_JMLR10}, the DL formulation in~\EEqref{eq:DL} is more difficult to solve because $\BF{M}'$ is dependent on $\zeta$.  
In~\cite{merlet_dictionary_MICCAI12} $\zeta$ and there $\BF{D}$ are iteratively updated using the Levenberg-Marquardt algorithm, which is actually problematic. 
Theorems~\ref{thm:optimalScale} and~\ref{thm:optimalDictionary} show that $\zeta$ should be determined adaptively from testing signals, not from training signals.

%

\begin{theorem}[Sparsity of Mixture of Tensors]\label{thm:mixture}
  Let $(\BF{D}_*, \zeta_*)$ be the optimal dictionary learned in~\EEqref{eq:DL_final} using signals generated from the single tensor model.  
 Let $\{\Vc_i'\}_{i=1}^p$ be the $p$ sparse coefficients under arbitrary given $p$ SPF coefficients $\{\Va_i'\}_{i=1}^p$ and $(\BF{D}_*,\zeta_*)$.  
 Let $\Vc_*'$ be the sparse vector corresponding to SPF coefficient $\Va'=\sum_{i=1}^p w_i \Va_i'$ and $(\BF{D}_*,\zeta_*)$, 
 with $\sum_i w_i=1$, $w_i\geq 0$. 
 Then we have $ \|\Vc_*'\|_1 \leq \max( \|\Vc_1'\|_1,\dots, \|\Vc_p'\|_1)$. 
\end{theorem}

      \begin{theorem}[Optimal Scale]\label{thm:optimalScale}
        If $\{\VV{e}_i\}$ is generated from the single tensor model with fixed mean diffusivity (MD) $d_0$, 
        then for large enough $N$, fixed $L$, and small enough $\epsilon_{\text{DL}}$, the optimal scale $\zeta$ for the DL problem in~\EEqref{eq:DL} is 
          $\zeta_* = (8\pi^2\tau d_0)^{-1}$.
        \end{theorem}
        
\begin{theorem}[Optimal Dictionary]\label{thm:optimalDictionary}
  For  signals generated from the single tensor model using a range of MD value $[d_0, t d_0 ]$, $t\geq 1$,  
  if the dictionary $\{\BF{D}_0, \zeta_0\}$ is the optimal solution for~\eqref{eq:DL_final}, 
  then for another range of $[d_1,t d_1]$,  $\{\BF{D}_0, \zeta_1\}$ is still optimal if $\zeta_0d_0=\zeta_1d_1$. 
\end{theorem}
The above theorems indicate that the optimal dictionary can be learned by using the single tensor model set with a range of MD values. The dictionary can 
then be applied adaptively to each voxel by adjusting the scale $\zeta$. 
In this work, we fixed $\zeta_0=(8\pi^2\tau d_0)^{-1}$, where $d_0=0.7\times 10^{-3}\,\text{mm}^2/\text{s}$, 
and generated the signals using the single tensor model with MD in range $[0.5,0.9]\times 10^{-3}$, FA in range $[0,0.9]$, 
and with the tensor orientated in $321$ directions equally distributed on $\mathbb{S}^2$. 
The corresponding SPF coefficients $\{\Va_j'\}$ in~\EEqref{eq:DL_final} were then computed with $N=4$, $L=8$ via numerical inner product. 
Efficient DL was then performed using the online method in~\cite{mairal_JMLR10} to learn $\BF{D}$, which is initialized using the identity matrix. 
By solving~\EEqref{eq:DL_final}, we learned $250$ atoms. 
After combined with the isotropic atoms $\{B_{n00}(\q)\}_{n=1}^N$, we have a total of $254$ atoms. 
Note that the isotropic atoms are important so that grey matter and the CSF can be sparsely represented; this is not considered in~\cite{merlet_dictionary_MICCAI12}. 
Given a testing signal vector $\VV{e}$, which represents a partial sampling of the $\q$-space, our method, called DL-SPFI for brevity, reconstructs the entire $\q$-space by first setting the scale $\zeta$ based on the estimated MD for the signal vector and then computing the signal-space coding coefficients $\Vc$ by solving
\begin{footnotesize}
\begin{equation}
  \min_{\Vc} \| \BF{M}'\BF{D}\Vc -\VV{e}' \|_2^2 +  \| \boldsymbol\Lambda \Vc \|_{1}
\end{equation}
\end{footnotesize}%
Note that additional regularization is imposed via $\boldsymbol\Lambda$, which is devised as a diagonal matrix with elements $\Lambda_{i}=\frac{S}{h_i}\lambda$, 
where $\lambda$ is the regularization tuning parameter, $S$ is the dimension of $\VV{e}'$, and $h_i$ is the energy of $i$-th atom, which essentially penalizes atoms with low energy. 
After estimating $\Vc$, the SPF coefficients $\Va=(\Va_0^T,\Va'^T)^T$ are obtained by first computing $\Va'=\BF{D}\Vc$ and then computing $\Va_0$ using~\EEqref{eq:a0}. 
Finally, the EAP/ODF can be obtained using closed-form expressions~\cite{Cheng_PDF_MICCAI2010,Cheng_ODF_MICCAI2010}.

\section{Experiments}
\label{sec:exp}

We compared the proposed DL-SPFI with $l_{1}$-FOCUSS (without DL) and DL-FOCUSS, both described in~\cite{bilgic_MRM12}. 
The DSI dataset and the codes provided by Bilgic\footnote{\href{http://web.mit.edu/berkin/www/software.html}{http://web.mit.edu/berkin/www/software.html}} were used. 
\medskip

\begin{figure}[t!]
\centering
\setlength{\abovecaptionskip}{2pt}
\setlength{\belowcaptionskip}{-15pt}
\begin{tabular}{c  c}
\begin{tikzpicture} 
  \begin{axis}
    [
      width=0.5\textwidth, 
      xlabel={FA (MD=$0.6\times 10^{-3}\,\text{mm}^2/\text{s}$, $\zeta_0$)}, ylabel={Number of Nonzero Coefficients}, 
      legend style={legend pos = north west, font=\tiny, legend cell align=left},
      xtick={0,0.1,0.2,...,0.9},
      ytick={0,20,...,180},
      grid=major, 
      xmin=0, xmax=0.9,
      ymin=0, ymax=180
    ] 
    
    \addplot[color=red,mark=square] 
    coordinates { (0, 2) (0.1,10.5639) (0.2,19.9252) (0.3,24.7259 ) (0.4,29.4611) (0.5,38.9657) (0.6,50.6199 ) (0.7,64.3364) (0.8,88.7103) (0.9,114.548) }; 
    \addlegendentry{SPF, single tensor}
    
    \addplot[color=red,mark=square*] 
    coordinates { (0, 2) (0.1,13.5483) (0.2,13.0187) (0.3,12.0748 ) (0.4,11.704) (0.5,11.8847) (0.6,12.3551) (0.7, 12.866) (0.8,13.2617) (0.9,14.4019) }; 
    \addlegendentry{DL-SPF, single tensor}
    
    \addplot[color=green!50!black,mark=square] 
    coordinates { (0, 2) (0.1,9.21495) (0.2,16.2648) (0.3,21.9688) (0.4,25.271) (0.5,31.0125) (0.6,40.9751 ) (0.7,53.1433) (0.8,71.6417) (0.9,102.059) }; 
    \addlegendentry{SPF, mixture of tensors}
    
    \addplot[color=green!50!black,mark=square*] 
    coordinates { (0, 2) (0.1,13.5514) (0.2,15.0779) (0.3,16.028 ) (0.4,16.6916) (0.5,17.1215) (0.6,17.271) (0.7, 18.6822) (0.8,19.6822) (0.9,20.0218) }; 
    \addlegendentry{DL-SPF, mixture of tensors}
    
  \end{axis} 
\end{tikzpicture}
&
\begin{tikzpicture} 
  \begin{axis}
    [
      width=0.5\textwidth, 
      xlabel={FA (MD=$1.1\times 10^{-3}\,\text{mm}^2/\text{s}$, mixture of tensors)}, ylabel={Number of Nonzero Coefficients}, 
      legend style={legend pos = north west, font=\tiny, legend cell align=left},
      xtick={0,0.1,0.2,...,0.9},
      ytick={0,20,...,180},
      grid=major, 
      xmin=0, xmax=0.9,
      ymin=0, ymax=180
    ] 
    
    

    \addplot[color=red,mark=square] 
    coordinates { (0, 4) (0.1,9.4891) (0.2,13.5981) (0.3,19.0249 ) (0.4,23.2617) (0.5,28.6106 ) (0.6,35.0249 ) (0.7,46) (0.8,65.9377) (0.9,94.2399) }; 
    \addlegendentry{SPF, $\zeta_0$}
    
    \addplot[color=red,mark=square*] 
    coordinates { (0, 3.99688) (0.1,8.93146) (0.2,15.8474) (0.3,21.6822) (0.4,26.0717) (0.5,29.6044) (0.6,40.9128) (0.7,70.6262) (0.8,108.093) (0.9,95.1963) }; 
    \addlegendentry{DL-SPF, $\zeta_0$}
    
    \addplot[color=green!50!black,mark=square] 
    coordinates { (0, 0) (0.1,11.2087) (0.2,16.4143) (0.3,18.5171 ) (0.4,22.3551) (0.5,29.785 ) (0.6,38.8287 ) (0.7,49.2617) (0.8,70.8349) (0.9,104.966) }; 
    \addlegendentry{SPF, adaptive $\zeta$}
    
    \addplot[color=green!50!black,mark=square*] 
    coordinates { (0, 0) (0.1,14.6231) (0.2,15.1153) (0.3,16.5296) (0.4,17.8505) (0.5,18.5826) (0.6,18.8567) (0.7,19.6199) (0.8,20.9502) (0.9,21.1059) }; 
    \addlegendentry{DL-SPF, adaptive $\zeta$}

  \end{axis} 
\end{tikzpicture}
\\
\begin{tikzpicture} 
  \begin{axis}
    [
      width=0.5\textwidth, 
      xlabel={Crossing Angle (SNR=$\infty$)}, ylabel={RMSE}, 
      legend style={legend pos = north west, font=\tiny, legend cell align=left},
      xtick={30,35,...,90},
      grid=major, 
      xmin=30, xmax=90,
      ytick={0,0.05,0.1,0.15,0.2,0.25},
      yticklabels={0,0.05,0.1,0.15,0.2,0.25},
      ymin=0, ymax=0.25
    ] 
   
    \addplot[color=blue,mark=square*] 
    coordinates { (30,0.132625) (35,0.131806) (40,0.122391) (45,0.11964)  (50,0.113717)  (55,0.108357) (60,0.103693) (65,0.102468) (70,0.107176) (75,0.103385) (80,0.106716) (85,0.111756) (90,0.112688)}; 
    \addlegendentry{DL-FOCUSS}
    
    \addplot[color=blue,mark=square] 
    coordinates { (30, 0.035385) (35,0.045658) (40,0.034387) (45,0.046714) (50,0.057274) (55,0.045792) (60,0.070762) (65,0.041414) (70, 0.040252) (75,0.047522)  (80,0.057808) (85,0.049412) (90,0.048333)}; 
    \addlegendentry{$l_{1}$-FOCUSS}

    \addplot[color=red,mark=square*] 
    coordinates {(30,0.0120366) (35,0.0114177) (40,0.0111471) (45,0.0121778) (50,0.0120073) (55,0.0124843) (60,0.0128742) (65,0.0134553) (70,0.0129282) (75,0.0128469) (80,0.0126444) (85,0.0117899) (90,0.0122857)};
    \addlegendentry{DL-SPFI}
    
    \addplot[color=red,mark=square]  
    coordinates {(30,0.0204835) (35,0.0213147) (40,0.0211575) (45,0.0238479) (50,0.0254104) (55,0.0228033) (60,0.0182591) (65,0.0271438) (70,0.031295) (75,0.0364104) (80,0.0287409) (85,0.0206489) (90,0.0134215)};
    \addlegendentry{$l_{1}$-SPFI}

  \end{axis} 
\end{tikzpicture}
&
\begin{tikzpicture} 
  \begin{axis}
    [
      width=0.5\textwidth, 
      xlabel={Crossing Angle (SNR=20)}, ylabel={RMSE}, 
      legend style={legend pos = south east, font=\tiny, legend cell align=left},
      xtick={30,35,...,90},
      grid=major, 
      xmin=30, xmax=90,
      ytick={0,0.05,0.1,0.15,0.2,0.25},
      yticklabels={0,0.05,0.1,0.15,0.2,0.25},
      ymin=0.0, ymax=0.25
    ] 
   
    \addplot[color=blue,mark=square*] 
    coordinates { (30,0.206978) (35,0.207226) (40,0.202141) (45,0.199469)  (50,0.200817)  (55,0.197007) (60,0.198238) (65,0.196964) (70,0.201001) (75,0.202912) (80,0.206188) (85,0.204468) (90,0.206462)}; 
    \addlegendentry{DL-FOCUSS}
    
    \addplot[color=blue,mark=square] 
    coordinates { (30, 0.199076) (35,0.201958) (40,0.203929) (45,0.205916) (50,0.207995) (55,0.210447) (60,0.211615) (65,0.212608) (70, 0.214577) (75,0.21614)  (80,0.215004) (85,0.214767) (90,0.216941)}; 
    \addlegendentry{$l_{1}$-FOCUSS}

    \addplot[color=red,mark=square*] 
    coordinates {(30,0.16864) (35,0.16774) (40,0.166329) (45,0.167901) (50,0.166765) (55,0.16273) (60,0.160195) (65,0.159542) (70,0.162979) (75,0.167667) (80,0.169452) (85,0.171443) (90,0.169408)};
    \addlegendentry{DL-SPFI}
    
    \addplot[color=red,mark=square]  
    coordinates {(30,0.232544) (35,0.229486) (40,0.22641) (45,0.227423) (50,0.226007) (55,0.22699) (60,0.226792) (65,0.226474) (70,0.230045) (75,0.231719) (80,0.229082) (85,0.227111) (90,0.223286)};
    \addlegendentry{$l_{1}$-SPFI}

  \end{axis} 
\end{tikzpicture}
\end{tabular}
\caption
{
  \footnotesize
  \label{fig:exp_synthetic}
  \textbf{Synthetic Experiments}. 
  The first row shows the average number of non-zero coefficients associated with the SPF basis and the DL-SPF basis for the single- and multiple-tensors and with and without adaptive scales. 
  The second row shows the RMSE values of various methods using the S\"oderman cylinder model with and without noise. 
}

\end{figure}
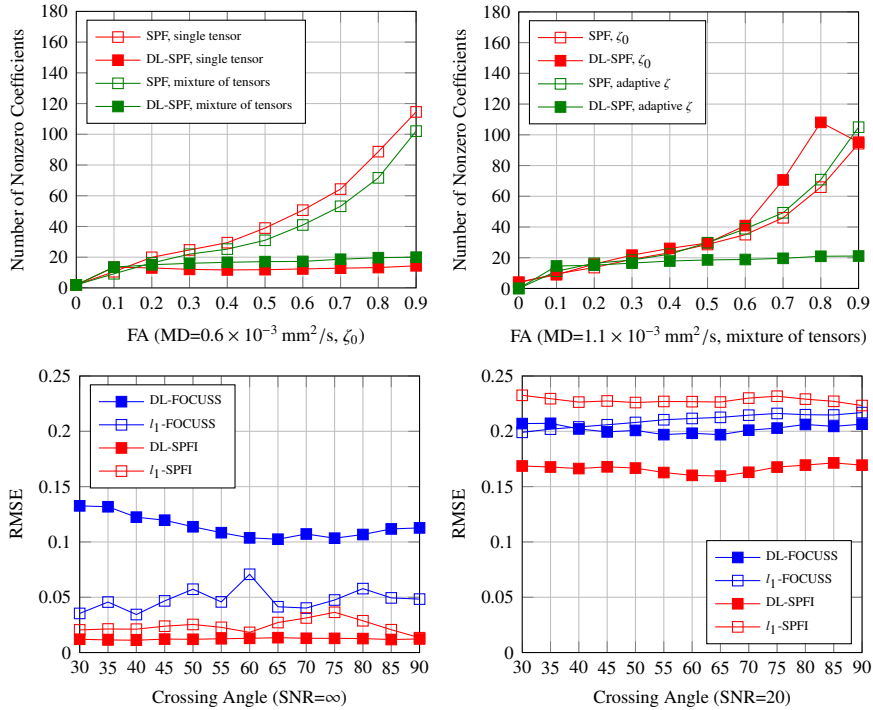

\noindent\textbf{Signal Sparsity}. 
The theorems were first validated. 
First, fixing MD value $d_1=0.6\times 10^{-3}\,\text{mm}^2/\text{s}$ and using FA range $[0,0.9]$, we generated sample signals from the single tensor model and the mixture of tensor model with the tensors orientated in random directions. 
The coefficients $\Va'$ for each sample were calculated via numerical integral with $N=4$, $L=8$, $\zeta=\zeta_0=(8\pi^2\tau d_0)^{-1}$. 
Coefficients $\Vc$ were then obtained via~\EEqref{eq:DL_final}, based on $\Va'$, $\zeta_0$, and the learned dictionary $\BF{D}$. 
The signal sparsity with respect to the SPF basis and the learned DL-SPF basis was then evaluated by counting the number of coefficients in $\Va'$ and $\Vc$ with absolute values larger than $0.01\|\Va'\|$ and $0.01\|\Vc\|$. 
The top left subfigure in Fig.~\ref{fig:exp_synthetic} shows that signal sparsity associated with the SPF basis decreases as FA increases, 
whereas sparsity for DL-SPF basis is quite consistent for both single- and multi-tensor samples, if MD value $d_1$ is within the MD range used during DL.  
This experiment validated Theorem~\ref{thm:mixture}. 
Second, we used MD value $d_2=1.1\times 10^{-3}\,\text{mm}^2/\text{s}$ and evaluated the signal sparsity of two bases associated with $\zeta_0$ and adaptive scale $\zeta=(8\pi^2\tau d_2)^{-1}$ for the mixture of tensor model using different FA values. 
The top right subfigure in Fig.~\ref{fig:exp_synthetic}
shows that even though the MD value $d_2$ of the testing signal is not within the MD range used in the training data, 
by adaptively setting the scale for the testing data, 
the signal can still be sparsely represented by the learned DL-SPF basis, thanks to theorem~\ref{thm:optimalDictionary}. 
Note that~\cite{merlet_dictionary_MICCAI12} learns the basis from randomly mixture of tensors, which is actually useless. 
Besides, in~\cite{merlet_dictionary_MICCAI12} the sparsity is tested using very limited number of samples with large noise, which is still problematic, 
because the sparsity should be tested in the original signal, not the estimated signal with large error due to overfitting. 

\medskip
\noindent\textbf{RMSE in Cylinder Model}. 
We also evaluated the DL-SPF basis using the S\"oderman cylinder model~\cite{OzarslanNeuroImage2006} different from the tensor model used in our DL process. 
Using the same DSI-based sampling scheme described in~\cite{bilgic_MRM12} ($b_{max}=8000 s/mm^2$, $514$ $\q$-space signal measurements), 
we generated a ground truth dataset using the cylinder model with the default parameters in~\cite{OzarslanNeuroImage2006}. 
Utilizing the evaluation method described~\cite{bilgic_MRM12}, we estimated the DL-SPF coefficients from the an under-sampled version of the ground truth dataset (generated using a power-law density function $R=3$~\cite{bilgic_MRM12}) and  
reconstructed the $\q$-space signals in the all $514$ directions. Reconstruction accuracy was evaluated with respect to the ground truth dataset using the root-mean-square error (RMSE). The same evaluations was repeated by adding Rician noise with signal-to-noise ratio (SNR) of $20$. For DL-SPFI and $l_{1}$-SPFI, we set $\lambda=\lambda_l=\lambda_n=10^{-8}$ for the noise-free dataset and $10^{-5}$ for the noisy dataset. 
The subfigures in the second row of Fig.~\ref{fig:exp_synthetic} indicates that 
DL-SPFI yields the lowest RMSE in both noiseless and noisy conditions, whereas the RMSE of $l_{1}$-SPFI increases significantly when the signal is noisy. 

\medskip
\noindent\textbf{RMSE in Real DSI Data}. 
We performed a similar evaluation using the real DSI data provided by Bilgic. 
In~\cite{bilgic_MRM12}, the dictionary of DL-FOCUSS was learned from one slice and reconstruction was performed on the other slice.  
The dictionary of DL-SPFI was learned from the synthetic signals and was applied directly to the testing slice. 
Due to noise, comparing the estimated signal to the real signal is not a proper way of evaluation. 
In~\cite{bilgic_MRM12}, the RMSE was computed with respect to a dataset with 10 averages, which is however not released. 
Therefore, we opted to report two types of RMSEs. 
The first is the RMSE ($2.82\%$) between the signal estimated from the under-sampled data with 170 samples and the signal estimated from the fully-sampled dataset with 514 samples. 
The second is the RMSE ($9.81\%$) between the signal estimated from under-sampled data and the measured signal. 
Note that the outcomes for these two types RMSEs are identical for DL-FOCUSS because the estimated signal from the fully-sampled dataset is simply the fully-sampled dataset itself due to the discrete representation. 
In this sense, the CR-DL is much more difficult than DR-DL. 
Since the first type of RMSE is small, we can conclude that, by using DL-SPFI, the under-sampled dataset is sufficient for reasonable EAP reconstruction. 
The scanning time can hence be significantly reduced. 

\newlength{\iwidth}
\newlength{\iheight}

\begin{figure}

\setlength{\iwidth}{0.1365\textheight}
\setlength{\iheight}{0.14\textheight}
\hspace{-5pt}
\centering
\begin{tabular}{p{1\iwidth}p{1\iwidth}p{1\iwidth}p{1\iwidth}p{0.05\iwidth}}
  
\begin{tikzpicture}[baseline,trim left]

  \draw (0\iwidth,0\iheight)[black, ultra thick] rectangle(1\iwidth,1\iheight);    
  \node[fancylabel, right=0pt, black, ultra thick] at (0\iwidth,1\iheight) { \ \ \textbf{\scriptsize{$l_{1}$-FOCUSS: 16.34\%}}};

  \begin{pgfonlayer}{background}
  \clip (0\iwidth,0\iheight) rectangle (1\iwidth,1\iheight);
  \node at (0.5\iwidth,0.5\iheight) {
    \includegraphics[width=\iwidth]{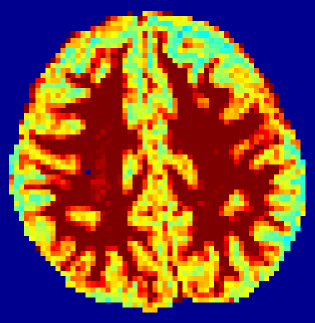}
    };
  \end{pgfonlayer}
\end{tikzpicture} 

&

\begin{tikzpicture}[baseline,trim left]

  \draw (0\iwidth,0\iheight)[black, ultra thick] rectangle(1\iwidth,1\iheight);    
  \node[fancylabel, right=0pt, black, ultra thick] at (0\iwidth,1\iheight) {\ \ \textbf{\scriptsize{DL-FOCUSS: 8.09\%}}};

  \begin{pgfonlayer}{background}
  \clip (0\iwidth,0\iheight) rectangle (1\iwidth,1\iheight);
  \node at (0.5\iwidth,0.5\iheight) {
  \includegraphics[width=\iwidth]{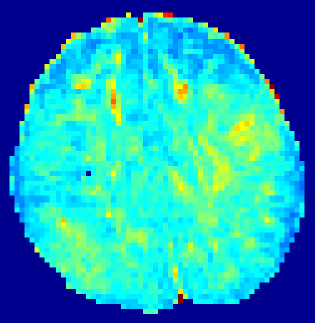}
  };
  \end{pgfonlayer}
\end{tikzpicture} 

&

\begin{tikzpicture}[baseline,trim left]

  \draw (0\iwidth,0\iheight)[black, ultra thick] rectangle(1\iwidth,1\iheight);    
  \node[fancylabel, right=0pt, black, ultra thick] at (0\iwidth,1\iheight) {\quad \textbf{\scriptsize{DL-SPFI: 2.82\%}}};

  \begin{pgfonlayer}{background}
  \clip (0\iwidth,0\iheight) rectangle (1\iwidth,1\iheight);
  \node at (0.5\iwidth,0.5\iheight) {
    \includegraphics[width=\iwidth]{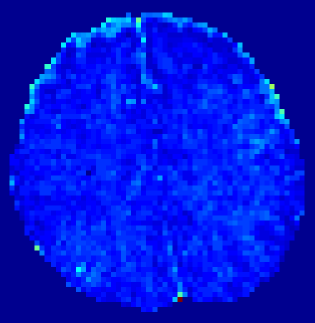}
    };
  \end{pgfonlayer}
\end{tikzpicture} 

&

\begin{tikzpicture}[baseline,trim left]

  \draw (0\iwidth,0\iheight)[black, ultra thick] rectangle(1\iwidth,1\iheight);    
  \node[fancylabel, right=0pt, black, ultra thick] at (0\iwidth,1\iheight) {\quad \textbf{\scriptsize{DL-SPFI: 9.81\%}}};

  \begin{pgfonlayer}{background}
  \clip (0\iwidth,0\iheight) rectangle (1\iwidth,1\iheight);
  \node at (0.5\iwidth,0.5\iheight) {
    \includegraphics[width=\iwidth]{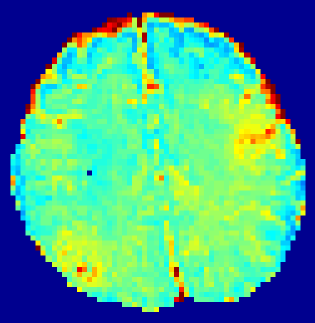}
    };
  \end{pgfonlayer}
\end{tikzpicture} 

&

\begin{tikzpicture}[baseline,trim left]

\setlength{\iwidth}{ 0.021\textheight}
\setlength{\iheight}{0.14\textheight}

  \begin{pgfonlayer}{background}
  \clip (0\iwidth,0\iheight) rectangle (1\iwidth,1\iheight);
  \node at (0.5\iwidth,0.5\iheight) {
    \includegraphics[width=\iwidth]{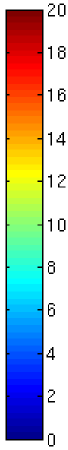}
    };
  \end{pgfonlayer}
\end{tikzpicture} 

\end{tabular}
\vspace{-20pt}
\caption{
  \footnotesize
\textbf{Real Data.} From left to right: the RMSE images for $l_{1}$-FOCUSS~\cite{bilgic_MRM12}, DL-FOCUSS~\cite{bilgic_MRM12},
  DL-SPFI between estimations using the under-sampled dataset and the fully-sampled dataset, 
  and DL-SPFI between the estimation from under-sampled data and the fully-sampled data. 
\label{fig:exp_real}}
\end{figure}
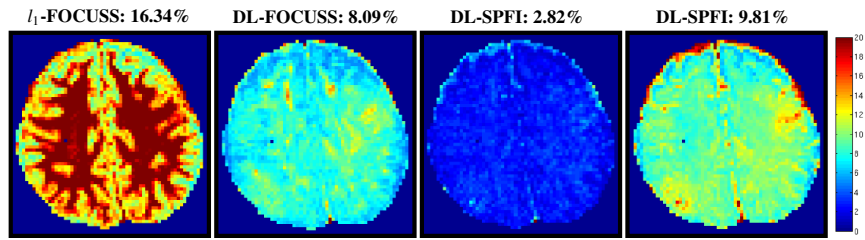

%
%

\section{Conclusion}
\label{sec:conclusion}

In this paper, we have demonstrated that DL-SPFI is capable of reconstructing the $\q$-space signal accurately using a reduced number of signal measurements. 
In DL-SPFI, an optimal dictionary is learned from exemplar diffusion signals generated from the single tensor model and can be applied adaptively to each voxel for effective signal reconstruction. 
Compared with the DR-DL based method in~\cite{bilgic_MRM12}, DL-SPFI avoids numerical errors by using a continuous representation of the $\q$-space signal, 
allowing the closed forms for EAP and the ODF estimation. 
Compared with the CR-DL method in~\cite{merlet_dictionary_MICCAI12}, DL-SPFI is significantly more efficient because the DL optimization is performed in a small dimensional subspace of the SPF coefficients, 
while DL in~\cite{merlet_dictionary_MICCAI12} is performed in a high dimensional space of fully sampled diffusion signal measurements in $\q$-space. 
Experimental results based on synthetic and real data indicate that DL-SPFI yields superb reconstruction accuracy using data with significantly reduced signal measurements.



\begin{footnotesize}
  \noindent \textbf{Acknowledgement}: This work was supported in part by a UNC start-up fund and NIH grants (EB006733, EB008374, EB009634, MH088520, AG041721, and MH100217). 
\end{footnotesize}



\end{document}